\documentclass[conference]{IEEEtran}
\IEEEoverridecommandlockouts
\usepackage{cite}
\usepackage{amsmath,amssymb,amsfonts}
\usepackage{algorithmic}
\usepackage{graphicx}
\usepackage{textcomp}
\usepackage{xcolor}
\def\BibTeX{{\rm B\kern-.05em{\sc i\kern-.025em b}\kern-.08em
    T\kern-.1667em\lower.7ex\hbox{E}\kern-.125emX}}
\begin{document}

\title{Graph-Augmented Cyclic Learning Framework for Similarity Estimation of Medical Clinical Notes\\
}


\author{Can Zheng, Yanshan Wang, Xiaowei Jia\\
University of Pittsburgh\\
\{caz51,yanshan.wang,xiaowei\}@pitt.edu}

\maketitle

\begin{abstract}
Semantic textual similarity (STS) in the clinical domain helps improve diagnostic efficiency and produce concise texts for downstream data mining tasks. However, given the high degree of domain knowledge involved in clinic text, it remains challenging for general language models to infer implicit medical relationships behind clinical sentences and output similarities correctly. In this paper, we present a graph-augmented cyclic learning framework for similarity estimation in the clinical domain. The framework can be conveniently implemented on a state-of-art backbone language model, and improve its performance by leveraging domain knowledge through co-training with an auxiliary graph convolution network (GCN) based network. We report the success of introducing domain knowledge in GCN and the co-training framework by improving the Bio-clinical BERT baseline by 16.3\% and 27.9\%, respectively.
\end{abstract}


\section{Introduction}
Electronic health records (EHRs) have been widely used in healthcare institutions to document patients' temporal medical conditions and actions, and showed great potential in improving clinical diagnostics, healthcare outcomes, and medical research\cite{blumenthal2011implementation}. However, studies \cite{wang2017characterizing} demonstrated the inefficiency and verbosity of clinical texts by showing the fact that in 23,630 progress notes written by 460 clinicians, 18\% was manually entered, 46\% was copied, and 36\% was imported. Those poorly organized and redundant texts are the results of the abuse of copy-and-paste, templates, and smart phrases \cite{embi2013computerized}. Besides, erroneous sentences are not uncommon in the free-text EHRs. Such non-customized and lengthy clinic records can add to a doctor’s burden of understanding and decline the efficiency as well as the quality of diagnosis processes.
Therefore, similar sentence removing techniques that are based on similarity estimation and return customized and concise clinic texts in the medical domain, are needed.\cite{kuhn2015clinical}.

General semantic textual similarity (STS) techniques have been developed in numerous natural language processing (NLP) applications such as text classification\cite{kowsari2019text} and topic detection\cite{makkonen2004simple}. In the clinical domain, the medical semantic textual similarity (MedSTS)~\cite{wang2020medsts} task provides a standard for evaluating relationships among text snippets, which is critical to downstream applications like clinical text abstraction, clinical semantics extraction, and clinical information retrieval. Especially, effective MedSTS helps  clarify the raw free-text EHRs by detecting and removing similar sentences. However, it remains a challenging task since clinical sentences often imply a lot of domain knowledge. Although prescription or medication information exists in structured EHRs, rich prescription information, such as the prescription from outside healthcare providers, is documented in the unstructured EHRs. These prescription descriptions are highly organized and context-poor. Only a professional doctor can distinguish them by potential diseases, prognosis, and drug relationships. Deep learning-based models like BERT~\cite{devlin2018bert} that are trained on semantically rich texts achieve huge success on general NLP tasks, but they often produce a sub-optimal performance for the MedSTS task~\cite{kades2021adapting}. Also, these methods often require a large amount of training data, whereas in the clinical domain there are always limited training data due to the sensitive patient information and time-consuming annotation.


\begin{figure}[htbp]
\centerline{\includegraphics[width=0.48\textwidth]{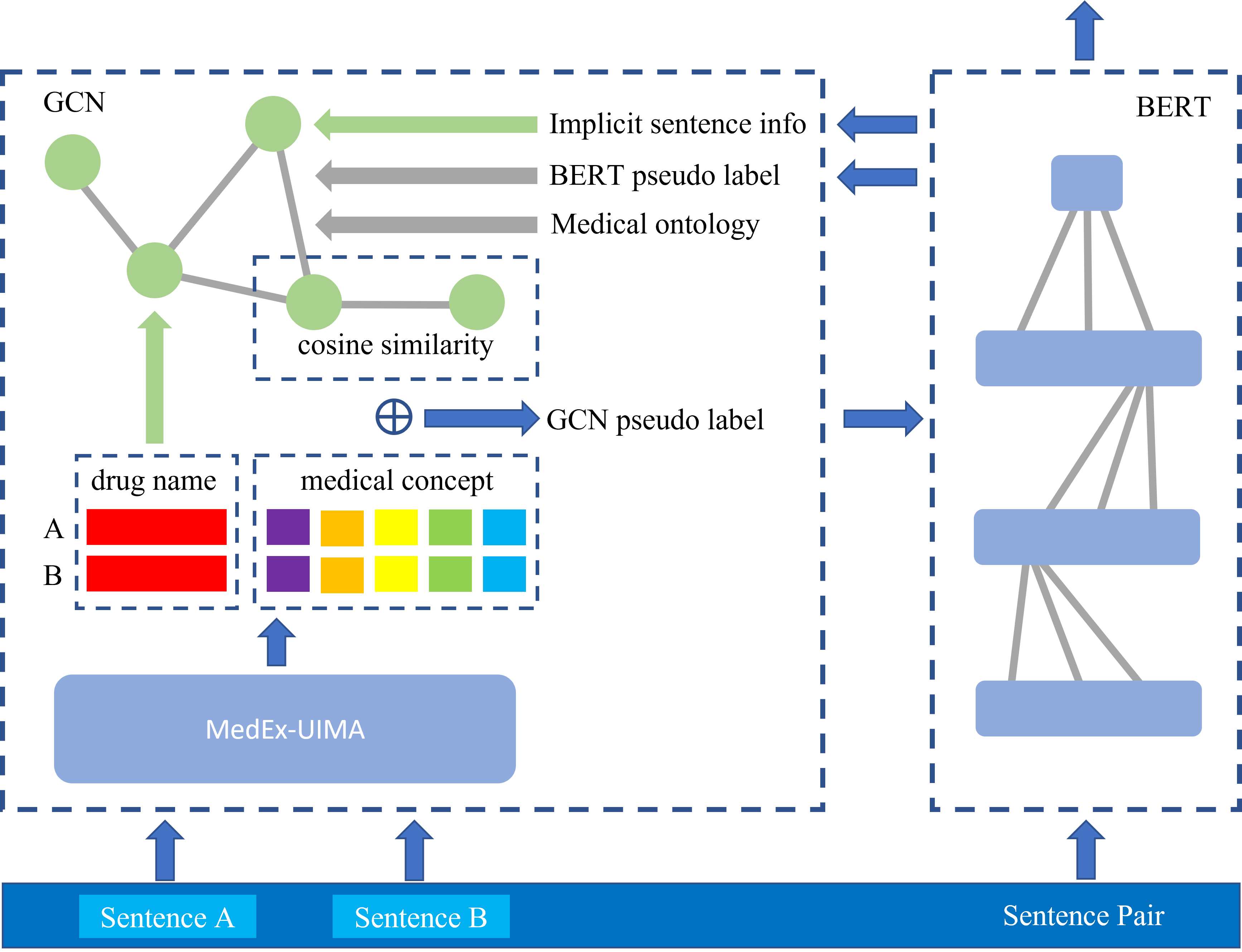}}
\caption{The structure of the cyclic generative framework is composed of a backbone network (BERT) and an assistant network (GCN). First, medical concepts of sentences in a given pair are extracted by MedEx-UIMA, including drug names and explicit info such as strength, frequency, etc. Nodes of the GCN align with drug names and are initially encoded by the BERT (green arrows). Drug-to-drug relationships in the graph are contributed by the ground truth labels, the medical ontology, and pseudo labels that are generated by the backbone network (gray arrows). After the completion of the GCN graph, the cyclic framework flows along with blue arrows: cosine similarity of drugs in sentence A and B is concatenated with the explicit feature, outputting GCN pseudo labels that will be fed to BERT in the next iteration; newly trained BERT then provide improved pseudo labels and help the graph construction again.
}
\label{frame+iter}
\end{figure}
In this paper, we proposed a cyclic generative knowledge-aware framework to boost deep learning-based language models for the MedSTS task on limited data by domain knowledge. The framework is able to augment the limited data by enlarging it with pseudo data and improve the performance of the backbone network by an extra auxiliary network that embeds medical domain knowledge. Basically, we designed a graph convolution network (GCN)~\cite{kipf2016semi} based auxiliary network to incorporate domain knowledge, which boosts the deep learning-based backbone network in a co-training style. Similarly, semantic information provided by the latter constrains the auxiliary GCN in graph construction. Our contribution can be summarized as follows:


\begin{itemize}
\item A generative co-training framework that augments clinical note data by pseudo data with increasing reliability and boosts all networks by iterations. In the cyclic process, the backbone network and auxiliary network annotate unlabeled data with different focus by turns, forming extra augmented data for training. Pseudo data that fits one network-specific criterion assist its following network. The goal is to leverage 
both the drug relationships from the GCN-based auxiliary network and the semantic information from the general language model when learning the sentence similarity using only limited annotated data.
\item A GCN-based auxiliary network enhanced by drug relationships. It can be locally constructed from the existing data set or an extra source of domain knowledge. By modifying the weights of edges in the graph with pseudo data from the backbone network, the auxiliary network emphasizes drug relationships with the semantic constraints, estimating unlabeled sentence pair more accurately. Also, drug-to-drug weights among the graph provide the interpretability of the model.
\end{itemize}

\section{Related work} 
Morden STS measurement pipelines are in two categories: a. ensemble models based on numerous sentence features, and b. pre-trained state-of-the-art models with task-specific fine tuning. For ensemble models, typical string-based similarity measurements are available, including Jaccard similarity\cite{niwattanakul2013using}, Q-gram similarity\cite{cavnar1994n}, tf-idf similarity\cite{salton1975vector}, etc. Those similarities in different semantic scales are then associated with encoding-based features that are generated by language encoders like Word2vec \cite{mikolov2013efficient}, InferSent \cite{conneau2017supervised}, and FastSent \cite{hill2016learning}, and finally predict the associated scores by random forest or dense neural networks~\cite{chen2018combining}. End-to-end deep learning-based models with fine tuning are the mainstream in this domain. State-of-the-art models like BERT~\cite{devlin2018bert} and its variants, and XLNet~\cite{yang2019xlnet}, are modified and involved in similarity calculating with task-specific fine tuning~\cite{wang20202019, kades2021adapting}. Also, multitask learning (MTL) paradigm is widely used and verified to learn powerful representations from multiple data\cite{zhang2017survey}. In the medical text similarity task, most top models involve MTL\cite{wang20202019,mahajan2020identification}. Despite the target similarity evaluation task, available tasks that boost models include sentence classification, medication named entity recognition, and sentence inference.

Being trained on large-scale corpora, contextual word embedding models like BERT encode more semantic information and dramatically improve performances for many fundamental NLP tasks\cite{devlin2018bert}. The basic BERT is composed of bidirectional transformers and is innovatively pre-trained on masked language model (MLM) and next sentence prediction tasks. Due to its powerful performance, many domain-specific versions have been produced: BioBERT\cite{lee2020biobert} and Bio-clinical BERT\cite{alsentzer2019publicly}. The bio-clinical BERT is one of our baselines and the backbone of the proposed framework, which is pre-trained on MIMIC-III and fine tuned on MedNLI and i2b2 named entity recognition (NER) tasks\cite{alsentzer2019publicly}. Therefore, it is suitable for evaluating the relationship between two sentences such as similarity. However, previous research\cite{kades2021adapting} showed that Bio-clinical BERT cannot handle well with prescriptions where less semantic information can be reflected. The author proposed a medication graph to enhance the backbone BERT model, but the graph is constructed from local data and cannot indicate relationships of unseen drug pairs objectively.


\section{Methods}
Our objective is to predict similarity score $y$ (ranging in [0,1,2,3,4,5]) given a pair of clinical note sentences $(a,b)$. Our data set $D = \{D_{tr}, D_{te}\}$ consists a training set $D_{tr}$ and a target testing set $D_{te}$. We have access to expert-annotated similarity scores for sentence pairs in the training set. Because the proposed method aims to analyze the similarity of existing clinical notes, we assume that we also have access to the target  sentence pairs in $D_{te}$ (but not labels) during the training process.  

Advanced machine learning or deep learning models trained on large volumes of text tend to use semantic similarity to assess sentence similarity, underplaying the role of drug relationships. To address this issue, we designed an auxiliary network that leverages drug relationships to continuously enhance the backbone network through a cyclic framework. As shown in Fig.~\ref{frame+iter}, the predictions produced by one network are used as pseudo data 
to augment the training data or the model structure of the other network. The goal is to leverage complementary strengths of two networks to capture both the semantic sentence-level similarity and drug-to-drug relationships through the cyclic learning process.

In this work, we adopt 
bio-clinical BERT~\cite{alsentzer2019publicly} as the backbone network in the proposed framework. The auxiliary network is designed to evaluate the similarity of a given sentence pair based on drug-to-drug relationships and medical concept differences. In particular, we create a GCN-based model to capture connections amongst different drugs and extract new drug representations that embed drug-to-drug relationships.  
Here we use GCN 
due to its ability in integrating global contextual information to enhance the drug embeddings. 
We will discuss the GCN model with two types of graphs,  
the local drug graph that is constructed by the local dataset, and the medical drug graph that is formed with the help of medical ontology. Besides the drugs in each sentence, the auxiliary network also considers the medical concept difference, e.g.,  
the difference in the strength and frequency of taking drugs in two prescription sentences. 
Such information can also be crucial in assessing sentence similarity, and thus we integrate the medical concept difference with the GCN model for the similarity estimation.  

In the following, we first introduce how to extract medical concept from clinical sentences, which will be used as the input to the auxiliary GCN network. Then we describe the auxiliary GCN networks with two types of structures, the local GCN structure and  the medical GCN structure. Finally, we summarize the overall cyclic learning process. 

\subsection{Medical concept extraction}

Drug names and other medical concept such as strengths, dose amount, route and frequency are critical for assessing the similarity of two sentences in the medical domain. 
The use of such information requires a mechanism to effectively extract these medical elements from original sentences and represent the difference in medical concept. 
In this paper, we apply a medical concept extraction tool, MedEx-UIMA~\cite{jiang2014extracting},  
which is an open-source semantic-based parser with unstructured information management architecture (UIMA) framework. MedEx-UIMA is used to find medical entities in a given sentence, including drug names, dose amount, route, frequency, etc.  

After obtaining multiple elements of each sentence, we will use the GCN model to represent the drug-to-drug relationships and also compute the difference 
of strength, unit type, frequency, tablet type, and dose, following the prior work~\cite{kades2021adapting}. For a numerical element $e_k$ ($k^{th}$ element) such as  frequency and dose, the difference is defined as the absolute residual: 
\begin{equation}
\Delta_k=\left | e_{k,1}-e_{k,2}\right |,
\end{equation}
where $e_{k,1}$ and $e_{k,2}$ represent the  $k^{th}$ element value in the first sentence and second sentence in the pair respectively. numerical element $k$ for a pair of sentences $S_1$ and $S_2$, respectively. 

For nominal variables $e_k$, the difference reflects the distinctness, and is represented as a binary number: 
\begin{equation}
\Delta_k=\left\{\begin{matrix}
0 & e_{k,1} = e_{k,2}\\ 
1 & e_{k,1} \neq  e_{k,2},
\end{matrix}\right.
\end{equation}

Based on these steps, we create the assemble feature ${d}_i=[\Delta_1, ..., \Delta_k, ..., \Delta_5]_i^T$ for each pair of sentences, which represents the difference in medical concept across two sentences. 

\subsection{GCN-based auxiliary networks}
\label{sec:gcn}
The standard deep learning-based backbone model often focuses on the semantic similarity between two sentences, but remains limited in capturing the drug relationships. 
For example, two sentences with similar sentence structures may have different meanings due to the difference in drugs. 
To fill in such gap, we build the GCN model as the auxiliary network to capture the drug relationships.

In particular, we create a graph $G(\mathcal{V},\mathcal{E},W)$, in which $V$ denotes the set of drugs, $\mathcal{E}$ denotes the set of edges connecting pairs of drugs,  and $W$ is the weighted adjacency matrix. 
Each node $v_i\in \mathcal{V}$ is associated with an initial embedding vector $x_i$, which is set as the  drug embedding produced by a pre-trained bio-clinical BERT model. 
Specifically, for a given drug and a sentence that contains it, we take 
drug embeddings 
from the last four layers' outputs of the pre-trained BERT, and then compute the average value as the drug embedding especially for this sentence. 
Then we set the initial embedding $x_i$ as the average drug embeddings of the drug $v_i$ over all the sentences that contain this drug. 
Our dataset contains 210 drugs, and thus we have the initial embedding matrix $X=[x_1, ..., x_n]^T$ in the shape of $210\times 768$, where $768$ is the embedding dimension. We consider two different methods to create the edge set $\mathcal{E}$ and the adjacency matrix $A$, and we call them as the local drug graph and the medical drug graph, respectively. 

\subsubsection{Local drug graph}
Here we first describe the local drug graph, which is built based on the available training clinical notes samples. 
Elements in the adjacent matrix $A$ are determined by the similarity scores from both expert-annotated training data and predictions made by the backbone model (i.e., the bio-clinical BERT model). 
We use $w_{ij}^g$ and $w_{ij}^p$ to represent the edge weight between a pair of drugs $v_i$ and $v_j$ estimated using ground-truth information (i.e., expert annotation) and pseudo labels (i.e., predictions by the backbone model). Specifically, the weight $w_{ij}^g$ is estimated as the average value of the annotated similarity scores $y_{ab}$ over all the training sentence pairs $(a,b)$ that have drugs $i$ and $j$. Similarly, the pseudo weight $w_{ij}^p$ is estimated as the average of the BERT-predicted similarity scores $\hat{y}_{ab}$ over all the training sentence pairs $(a,b)$ with drugs $i$ and $j$. Finally, the edge weight $w_{ij}$ is estimated through a linear combination of $w_{ij}^g$ and $w_{ij}^p$. More formally, the estimation of adjacency weights is shown as follows:
\begin{equation}
\begin{aligned}
w_{ij}^g&=\frac{\sum_{(a,b)\in D_{tr}:(i,j)}{y_{ab}}}{n_{g}},\\
w_{ij}^p&=\frac{\sum_{(a,b)\in D:(i,j)}{\hat{y}_{ab}}}{n_{p}},\\
w_{ij}&=(1-\alpha)w_{ij}^g+\alpha w_{ij}^p,
\end{aligned}
\end{equation}
where $(a,b)\in D:(i,j)$ indicates any sentence pairs $(a,b)$ in the dataset $D$ that contains the drugs $i$ and $j$, $n_{g}$ is the number of sentence pairs that in the training set and contain drug $i$ as well as drug $j$, $n_{p}$ is the number of such sentence pairs in both the training and target testing data, and the hyper parameter $\alpha$ is introduced to balance the contribution from the two sources. 
If 
$n_T$ is equal to 0, the weight $w_{ij}^g$ will be ignored and $w_{ij}$ will only be determined by $w_{ij}^p$. When both $n_g$ and $n_p$ are equal to 0, i.e., drugs $i$ and $j$ never appear in any sentence pairs in the data set, we will not include this edge $(i,j)$ in the edge set $\mathcal{E}$.  


\subsubsection{Medical drug graph}
According to the previous discussion, the local drug structure does not include edges between drugs $(i,j)$ if there exist no sentence pairs in the data set that contains both of them.   
Given the limited size of the available data set, the local graph can be sparse and incomplete. To fill up the graph and augment the local graph, 
we design the medical drug graph by leveraging the medical information. The weights of the medical drug graph ($\hat{w}_{ij}$) is composed of that of local drug graph ($w_{ij}$) and medical weights ($w_{ij}^o$).

The medical weights $w_{ij}^o$ are used to quantify  the medical relationship of two drugs based on  
the RxNORM ontology~\cite{RxNORM}. 
It provides normalized names for clinical drugs and their corresponding generic ingredients as CUIs. RxNORM ontology links 106,791 nodes with 51 properties. In this paper, we used the property of \textit{ingredient} to form the medication graph. The medical weight of drug $i$ and $j$ is defined based on the  shortest distance between drug $i$ and $j$ on the drug ontology (represented as $p_{ij}$),  as follows:
\begin{equation}
\begin{aligned}
w_{ij}^o&=\frac{5(p_{max}-p_{ij})}{p_{max}}, \\
p_{ij}&=p_{max}, \text{ if } p_{ij} \text{ does not exist},
\end{aligned}
\end{equation}
where 
$p_{max}$ is the longest path distance among all drug pairs in the data set. The medication adjacency level is inversely proportional to the path distance. We also transform the medical weights by scaling the adjacency to the range of $[0, 5]$, which fits the score range in the data set and stays consistent with the scale of weights in the local drug graph. After obtaining the medical weights, an overall weight is generated combining the annotated similarity scores from the data set and the medical ontology, as 
\begin{equation}
\hat{w}_{ij}=(1-\alpha)w_{ij}^g+\alpha ((1-\beta )w_{ij}^p+\beta w_{ij}^o). 
\end{equation}

Here $w_{ij}^o$ is considered as another type of pseudo weight, and thus share the contribution factor $\alpha$ with $w_{ij}^p$.  Another hyper-parameter $\beta$ is introduced to balance the competition between domain knowledge 
and the pseudo weight in the local graph. In the absence of $w_{ij}^g$, $\alpha $ is set to $1$, supposing the relationship between two drugs fully relies on the backbone model predictions and medical ontology.

\subsection{Enhancing Backbone Model Using Auxiliary Network}
We use the GCN model to embed each drug by incorporating its relationships with other drugs using the constructed graph structure. In particular, for each drug $i$, the GCN model outputs a hidden representation at each layer $L$ by aggregating the information from drug $i$'s neighborhood $N(i)$, as 
\begin{equation}
    h^L_i = \sum_{j\in N(i)} \text{tanh}(Uh^{L-1}_j+v),
\end{equation}
where $U$ and $v$ are model parameters. In this work, we set the GCN output representation in a shape of $128\times 1$. For a sentence pair, cosine similarity of the corresponding drugs is then calculated and concatenated with the assemble difference features ${d}_i$ of other medical concept. This concatenated vector is then fed into another regression layer to produce the output of the auxiliary network. 

Our proposed cyclic framework is composed of the  backbone network (i.e., bio-clinical BERT model) and the GCN-based auxiliary network. The cyclic learning process iteratively uses the predictions of one network 
to improve the training of the other network. 
In particular, the predictions of the backbone network will be used as the pseudo weights for creating the graph used by the auxiliary GCN model (as discussed in Section~\ref{sec:gcn}). Moreover, the outputs from one model are used as pseudo labels to regularize the training objective of the other model. We use the mean square error (MSE) loss $\mathcal{L}$ for both  models. The training objective functions for the backbone network ($\mathcal{L}_B$) and the auxiliary network ($\mathcal{L}_A$) combine the annotated labels in training data $D_{tr}$ and the pseudo labels in the target data $D_{te}$, as 
\begin{equation}
\footnotesize
\begin{aligned}
    \mathcal{L}_{B} = \mathcal{L}_{((a,b),y)\in D_{tr}}[f_{B}(a,b),y] + \gamma_B \mathcal{L}_{(a,b)\in D_{te}}[f_{B}(a,b),f_{A}(a,b)], \\
    \mathcal{L}_{A} = \mathcal{L}_{((a,b),y)\in D_{tr}}[f_{A}(a,b),y] + \gamma_A \mathcal{L}_{(a,b)\in D_{te}}[f_{B}(a,b),f_A(a,b)], 
\end{aligned}
\end{equation}
where $f_{B}$ and $f_{A}$ represent the transformation defined by the backbone model and the auxiliary model, respectively.  We introduce hyper-parameters $\gamma_A$ and $\gamma_B$ to control the contribution of pseudo labels. A larger value of $\gamma_A$ and $\gamma_B$ transfers more information from pseudo labels while also being impacted by inaccurate predictions in pseudo labels.  In our test, we set $\gamma_B$ as 0.5, and we found that setting $\gamma_B$ between [0,1] can always help improve the training of the backbone model. On the other hand, the value of $\gamma_A$ is found to have limited impacts on the learning of the auxiliary networks because the pseudo labels produced by the backbone model have already been incorporated in creating the graphs in the auxiliary model.    


\section{Experiment}
\subsection{Data Set}

\begin{figure}[htbp]
\centerline{\includegraphics[width=0.5\textwidth]{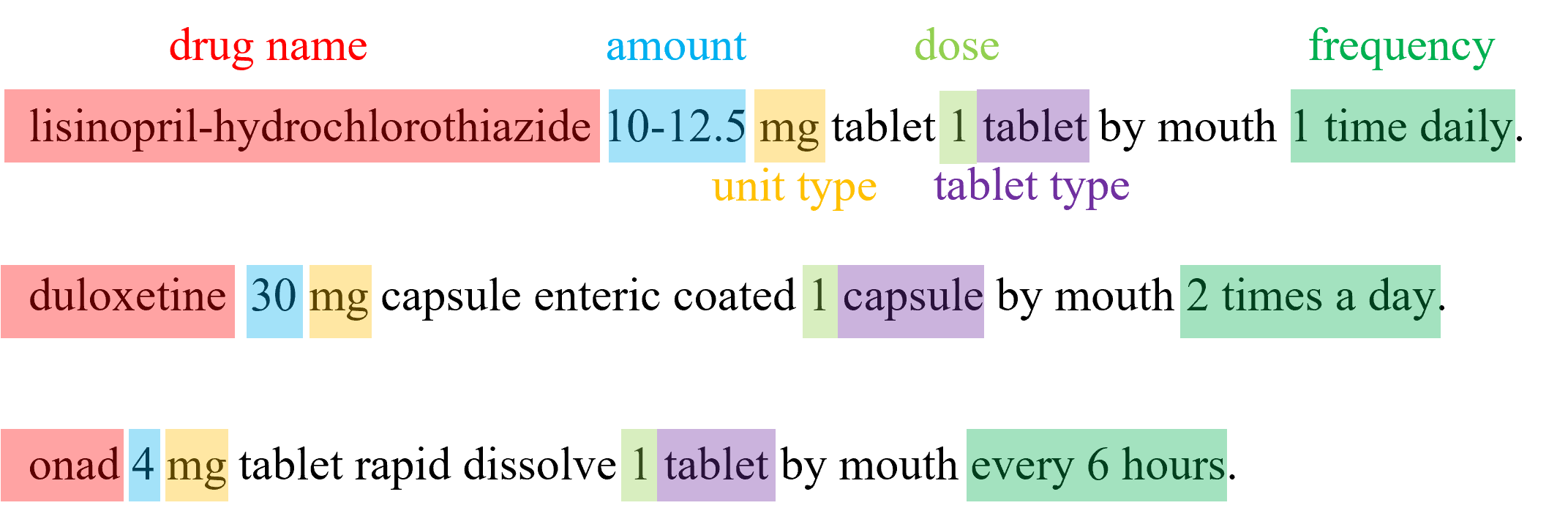}}
\caption{Some examples of tablet sentences and medical concept that are used in this paper.}
\label{senexample}
\end{figure}

Dataset used in this paper is a subset of the 2019 n2c2/OHNLP ClinicalSTS dataset ~\cite{wang20202019}. The comprehensive ClinicalSTS data set labels 2054 real-word clinical text pairs with similarity scores ranging from 0 to 5. According to the previous research, end-to-end models are hard to evaluate the similarity of tablet sentences that are highly related to domain knowledge and have limited semantic information (Fig.\ref{senexample}).
Therefore, we used MedEx-UIMA tool to select tablet sentences that at least has drug name and strength. In total 397 tablet sentence pairs are collected, forming the subset data set for this paper. Then the training set and testing set are split by 65:35 from the subset data.

\subsection{Experiment settings}
We compare to multiple baseline models 
in this paper. We first compared to the Bio-clinical BERT that was fine-tuned on our data set. Graph-related model in previous work~\cite{kades2021adapting} was also tested on the data set to validate the improvement provided by extra medical knowledge. To prove that the cyclic framework better boosts the backbone network as well as the assistant network, we designed a simple ensemble model (Ensemble) that outputs the similarity score of a sentence pair by the weighted sum of BERT result and medical GCN result. 
We reported two versions of our model, cyclic framework embedded with BERT and GCN on the local graph (BERT+local GCN), and cyclic framework embedded with BERT and GCN on the medical graph (BERT+medical GCN). 

Drug relationships for constructing the graph structure are quantified by averaged expert annotated sentence similarity scores ($w^g$), averaged pseudo sentence similarity scores ($w^p$), and corresponding medical scores ($w^o$). In this paper, we regard $w^p$ and $w^m$ weights that assist the solid ground truth weight. Therefore, it is necessary to explore the effect of the contribution ratio of $w^p$ and $w^m$, which is $\beta$ on the final performance. This helps to demonstrate the improvement of the medical graph over the local graph when $\beta$ increases.

\section{Results}
We use the Pearson correlation coefficient to evaluate our models. 
It is worthwhile to mention that the model is trained by minimizing the MSE loss, which is different from the correlation measure used here. Results are summarized in Table~\ref{result}. The BERT baseline in this limited data set results in a correlation of 0.43 with 200 training epochs. local GCN improves the overall performance by 0.07, and the medical GCN boosts it further to 0.55. The graph network that aggregates drug relationships (Graph-related model) outperforms the backbone BERT by at least 0.04. Comparing the Ensemble models with GCNs and corresponding cyclic models, we see that the proposed framework is more efficient in boosting two networks in iterations than simple ensemble learning. In the following, we will study the proposed method from three aspects: (i) the interaction between BERT and GCN, (ii) the comparison between the local GCN and the medical GCN, and (iii) the effect of leveraging ontology information.
\begin{table}[htbp]
\caption{Pearson correlation coefficient of different models}
\begin{center}
\begin{tabular}{lc}
\hline
\textbf{Model} & \textbf{Correlation}\\
\hline
Bio-clinical BERT & {0.43} \\
Graph-related model~\cite{kades2021adapting} & {0.47} \\
Ensemble (local GCN) & {0.46}\\
Ensemble (medical GCN) & {0.50}\\\hline
BERT+local GCN & {0.50} \\
BERT+medical GCN & {0.55}  \\\hline
\end{tabular}
\label{result}
\end{center}
\end{table}

\begin{figure}[htbp]
\centerline{\includegraphics[width=0.48\textwidth]{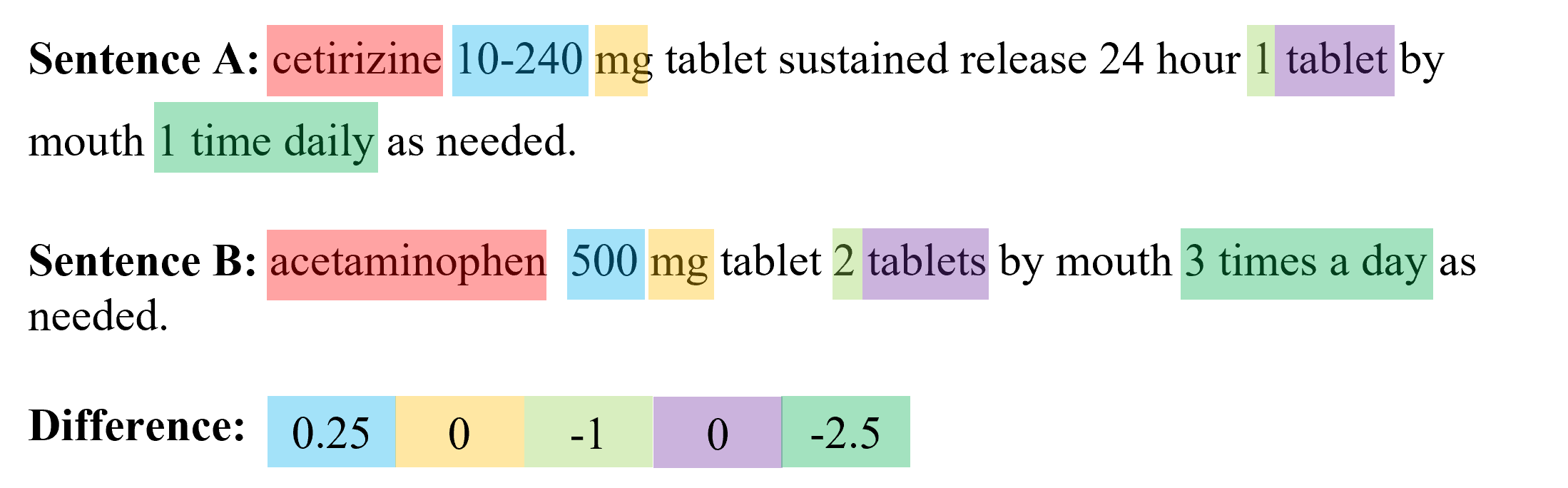}}
\caption{An example that is overestimated by medical GCN but well predicted by BERT.}
\label{leftexample}
\end{figure}
\subsection{BERT vs. GCN}
\begin{figure}[htbp]
\centerline{\includegraphics[width=0.48\textwidth]{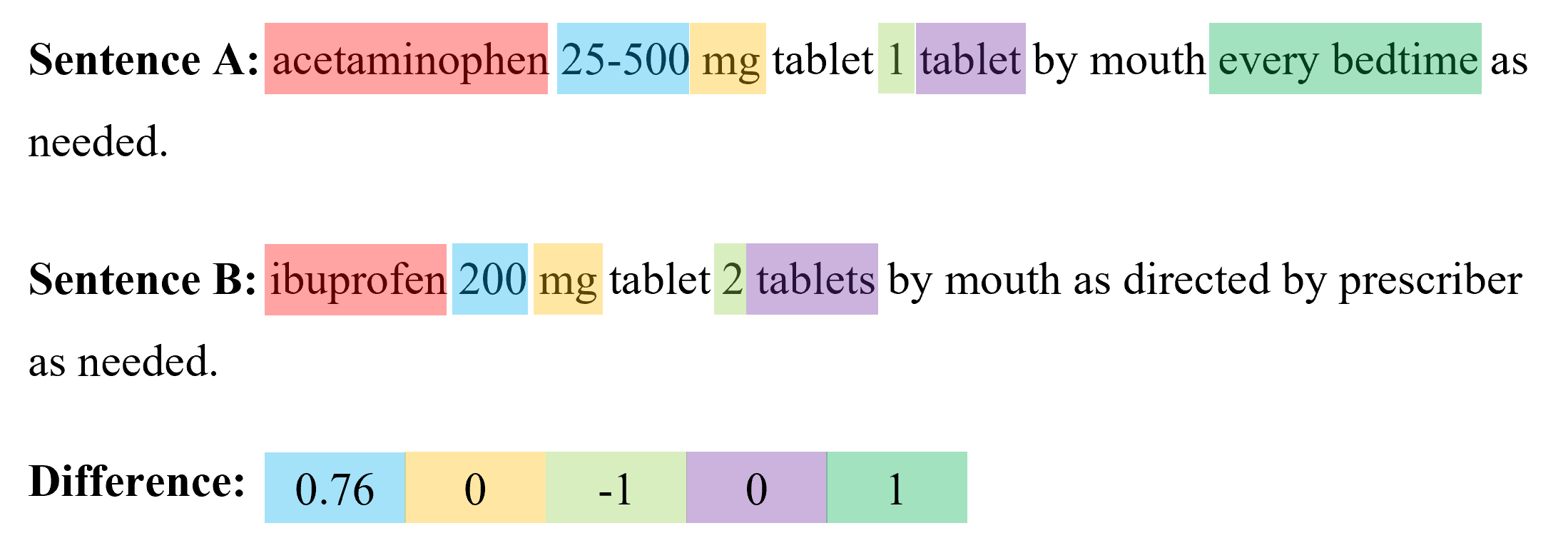}}
\caption{An example that is overestimated by BERT but well predicted by medical GCN.}
\label{rightexample}
\end{figure}

\begin{figure*}[htbp]
\centerline{\includegraphics[width=1\textwidth]{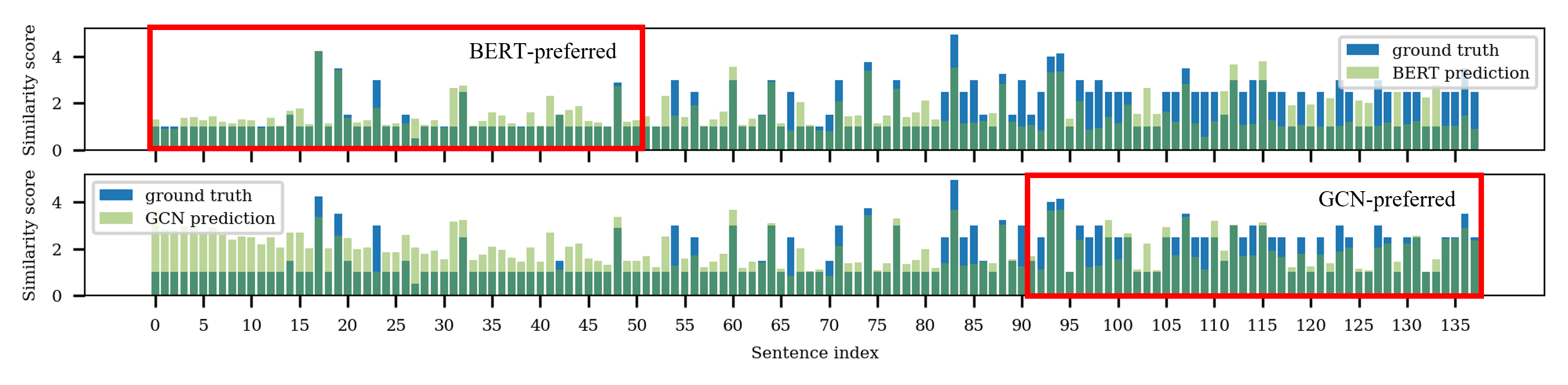}}
\caption{Comparisons of predicted similarity scores generated by BERT and the ground truth similarity scores (upper), and that of predicted similarity scores generated by medical GCN and the ground truth similarity scores (bottom). Accurate predictions are displayed as well-overlapping bars in deep green.}
\label{detailed_result}
\end{figure*}

Figure \ref{detailed_result} shows the comparison among ground truth, and predictions of BERT and medical GCN in the first iteration. The horizontal axis is sorted by the difference of distance between BERT's prediction and the true value and that between GCN's prediction and the true value, so BERT performs better on the sentences indexed on the left side of the plot (1 to 50), while GCN does better on the sentences indexed on the right side of the graph (91 to 138). For sentence pairs that are indexed from 51 to 90, the backbone model and auxiliary model perform similarly. Starting from the left side, BERT-preferred sentence pairs are overestimated by the medical GCN model since most of them contain related drugs while are less similar in terms of symmetric meanings. For the sentence pair shown in Figure \ref{leftexample}, the difference of amount, frequency, and dose indicates different illnesses. But two drugs are highly related in the medical ontology, misleading the GCN to believe that the two sentences are for the same condition. 
In fact, the shortest path from "cetirizine" to "acetaminophen" is 4 since both drugs are usually combined with "pseudoephedrine", forming polypills. Meanwhile, the GCN helps to right value sentence pairs that have strong medical relationships and are underestimated by the BERT. An example is shown in Figure \ref{rightexample}. Given the similar amount and related drugs, the GCN correctly scores the similarity between sentences A and B. The shortest path of "acetaminophen" between "ibuprofen" is 2, which is also validated by the truth that they are non-steroidal antipyretic and analgesic drugs that relieve fever and pain by inhibiting the synthesis of prostaglandin. For the rest pairs, the auxiliary model behaved as well as or as badly as the backbone model. 
\subsection{Local GCN vs. medical GCN}
\begin{figure}[!t]
\centerline{\includegraphics[width=0.45\textwidth]{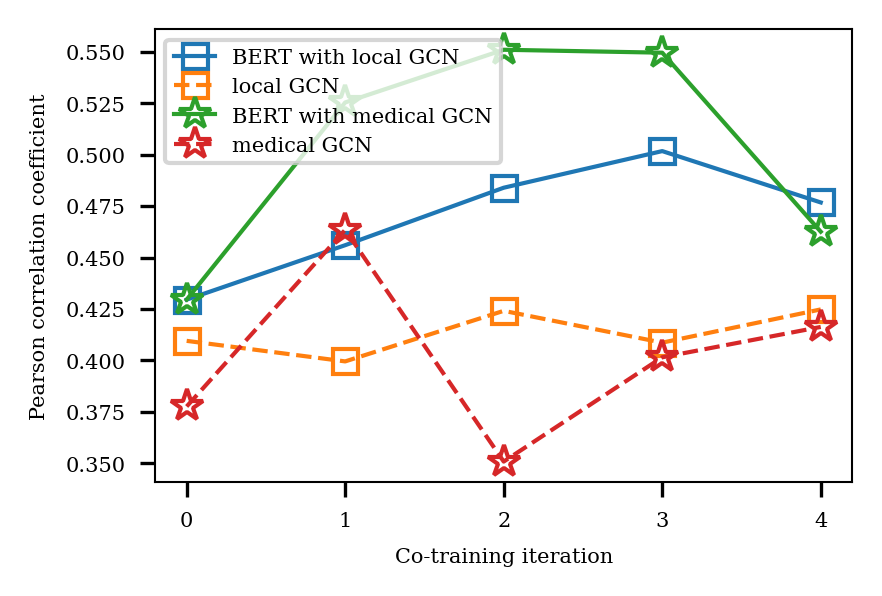}}
\caption{Pearson correlation coefficients of all sub networks. $\alpha = 0.5, \beta = 0.5$ for medical GCN, and $\alpha = 0.2, \beta = 0$ for local GCN.}
\label{localvsmed}
\end{figure}

Figure~\ref{localvsmed} shows the performance of sub-networks in terms of the Pearson correlation coefficient during co-training iterations. $\alpha=0.2$, $\beta=0$ for local GCN and $\alpha=0.5$, $\beta=0.5$ for medical GCN. We can see that starting from the same initial BERT, the enhanced BERT with medical GCN is superior to that with local GCN. Correlation coefficients of both backbone networks achieve peaks at the third iteration and then drop to the same level around 0.47 finally. However, the performance of GCNs is not correlated with that of their backbone networks, especially at the third and fourth iterations where correlation coefficients of backbone network achieve their maximums while that of GCNs heavily drop. A possible reason is that with the increasing accuracy of BERT predictions, more semantic information is involved, but GCN that focuses on drug relationships is not able to provide such comprehensive representations either boost the backbone model. It also leads to the eventual degradation of the backbone network performance. Besides, GCN is not as stable as BERT on our small dataset.

\subsection{Ontology contribution}
\begin{figure}[!t]
\centerline{\includegraphics[width=0.45\textwidth]{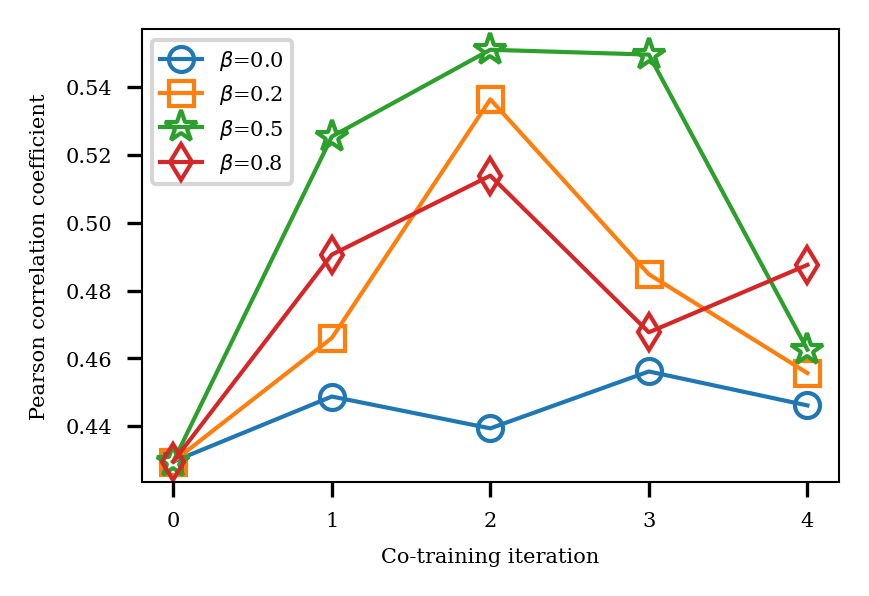}}
\caption{Pearson correlation coefficients of the backbone network (BERT) with different partitions of ontology information, $\alpha=0.5$.}
\label{onto}
\end{figure}

To explore the effect of extra medical knowledge, we applied different $\beta$ values from $0$ to $1$. In this part, $\alpha$ is set to $0.5$. According to Figure \ref{onto}, the performance of the backbone network is not proportional to the contribution of ontology knowledge. Less involvement of medical knowledge $(\alpha=0.5, \beta=0.2)$ heavily boost the baseline model that no extra knowledge is attended $(\alpha=0.5, \beta=0.0)$. But the performance degrades when pseudo scores are dominated by ontology information because ground truth similarities that contain semantic information are not effectively combined with ontology weights by simple weighted sums. A possible reason is that the medical graph is locally comprehensive, and zero elements in the adjacency matrix are meaningful, indicating that the two drugs are independent. However, when pseudo labels are introduced, those elements will be filled by non-zero values, even though they are small, and the following normalization will wrongly aggregate these pseudo weights, causing the graph to misrepresent drug relationships. Configuration $(\alpha=0.5, \beta=0.5)$ that balances ground truth as well as medical ontology best boost the performance of the backbone network.

\section{Conclusion}
In this paper, we proposed a graph-augmented cyclic learning framework for similarity estimation of medical clinical notes, which is composed of a backbone model and an auxiliary medical GCN. The framework is able to enlarge limited data set by annotating unlabeled data and successfully boost the backbone model. Meanwhile, such a cyclic structure can integrate the information from the auxiliary network, providing an approach to leverage extra knowledge. The GCN-based auxiliary networks effectively encode domain knowledge and transfer it to the backbone network; the backbone network also maintains the semantic accuracy for the former. The system is finally boosted to a Pearson correlation coefficient of 0.55, which is increased by 27.9\% of the origin backbone model result. In addition, GCN provides the model interpretability to a certain extent, splitting the similarity score into drug similarity and sentence similarity. Also, we discussed the effect of three super parameters in this framework, which inspires further potential improvements.

Future work can focus on a more compact combination of the auxiliary network and the backbone model, in terms of methods that enhance each other and GCN construction. According to our experiments, the improvement achieved by adjusting $\gamma_A$ and $\gamma_B$ is not obvious. For GCN construction, normalization of the sparse adjacent matrix results in the failure of manual labels and ontology scores to combine. It would further improve the performance of the GCN by correctly introducing task-specific labels. Besides, larger data sets help further evaluate the proposed framework and stabilize the GCN.



\bibliographystyle{IEEEtran}
\bibliography{IEEEabrv}

\end{document}